\documentclass[sigconf]{acmart}

\usepackage{algorithm}
\usepackage[noend]{algpseudocode}
\usepackage{booktabs} % For formal tables

\newcommand*{\affaddr}[1]{#1} 
\newcommand*{\affmark}[1][*]{\textsuperscript{#1}}
\AtBeginDocument{%
  \providecommand\BibTeX{{%
    \normalfont B\kern-0.5em{\scshape i\kern-0.25em b}\kern-0.8em\TeX}}}

\acmYear{}
\acmPrice{}
\acmISBN{}
\acmDOI{}
\setcopyright{none}
\acmConference[AdvML'19: Workshop on Adversarial Learning Methods for Machine Learning and Data Mining at KDD]{AdvML'19: Workshop on Adversarial Learning Methods for Machine Learning and Data Mining at KDD}{August 5th, 2019}{Anchorage, Alaska, USA}

\begin{document}
\title[Localized Adversarial Training]{Localized Adversarial Training for Increased Accuracy and Robustness in Image Classification}


%\author{Eitan Rothberg}
%\email{rothberg.10@osu.edu}
%\affiliation{%
%  \institution{Ohio State University}
 % \city{Columbus}
 % \postcode{43210}
%}
%\author{Tingting Chen}
%\email{tingtingchen@cpp.edu}
%\author{Hao Ji}
%\email{hji@cpp.edu}
%\affiliation{%
%  \institution{California State Polytechnic University}
 % \city{Pomona}
 % \state{California}
 % \postcode{91768}
%}
%\author{Luo Jie}
%\email{roger@sigmoidai.com}
%\affiliation{%
%   \institution{Sigmoid AI}
%  \city{San Francisco Bay Area}
 % \state{California}
%   \postcode{??}
%}

\author{Eitan Rothberg\affmark[1], Tingting Chen\affmark[2], Luo Jie\affmark[3], and Hao Ji\affmark[2]\\}
\affiliation{\affaddr{\affmark[1]{Ohio State University, Columbus, Ohio, 43210, USA\\}}
\affaddr{\affmark[2]California State Polytechnic University, Pomona, CA, 91768, USA\\}
\affmark[1]rothberg.10@osu.edu, \affmark[2]\{tingtingchen, hji\}@cpp.edu, \affmark[3]roger@sigmoidai.com\\}

\renewcommand{\shortauthors}{Rothberg, et al.}

\begin{abstract}
    Today's state-of-the-art image classifiers fail to correctly classify carefully manipulated adversarial images. In this work, we develop a new, localized adversarial attack that generates adversarial examples by imperceptibly altering the backgrounds of normal images. We first use this attack to highlight the unnecessary sensitivity of neural networks to changes in the background of an image, then use it as part of a new training technique: localized adversarial training. By including locally adversarial images in the training set, we are able to create a classifier that suffers less loss than a non-adversarially trained counterpart model on both natural and adversarial inputs. The evaluation of our localized adversarial training algorithm on MNIST and CIFAR-10 datasets shows decreased accuracy loss on natural images, and increased robustness against adversarial inputs. 
\end{abstract}

\maketitle

\section{Introduction}
Since the advent of machine learning to the field of computer vision, image classification software has surpassed human capabilities and enabled new technologies including facial recognition authentication, self-driving cars, and smart security cameras \cite{DBLP:journals/corr/abs-1801-00553}. However, a unique challenge threatens these technologies: the existence of images which appear normal to humans, but reliably fool image classifiers \cite{szegedy_zaremba_sutskever_bruna_erhan_goodfellow_fergus_2014}. Because convolutional neural networks (CNNs) tend to focus on minor and easily manipulated details, attacks such as the Fast Gradient Sign Method (FGSM) \cite{goodfellow_shlens_szegedy_2015}, and its iterative counterpart, Projected Gradient Descent (PGD)  \cite{kurakin_goodfellow_bengio_2017}, have been able to reliably generate such adversarial examples by reverse engineering the training process. In the same way a classifier's weights are updated to minimize its loss during training, adversarial attacks seek to maximize a classifiers' loss by carefully altering images. 

Adversarial examples are a double-edged sword. On the one hand, they provide a unique angle for studying the learning process of neural networks. Because they allow us to follow the progression of an image as it changes from one class to another, adversarial examples can be helpful for understanding and interpreting decision boundaries. This information can be used for building more accurate classifiers, and might enable model-based optimization  \cite{adversarial:ccs16}.

On the other hand, adversarial examples pose a potentially existential threat to the safety of vision-dependent technologies. If image classifiers can be fooled, then so can the technologies that use them. Adversarial attacks have already been experimented with against self-driving cars and facial recognition authentication. \\\indent Adversarial training, the process of including adversarial examples in the training set, is one popular defense \cite{goodfellow_shlens_szegedy_2015}. However, while this technique has been shown to improve the robustness of a classifier against adversarial examples, the inclusion of adversarial images in the training process weakens classifiers' accuracy on natural (unaltered) images \cite{tsipras_santurkar_engstrom_turner_madry_2018} \cite{su_zhang_chen_yi_chen_gao_2018}. \textit{Building classifiers that maintain state-of-the-art accuracy on both natural and adversarial examples is a key challenge in image classification}, as a solution would provide defense as well as insight into the nature of CNNs \cite{tsipras_santurkar_engstrom_turner_madry_2018}. \\\indent We present a simple solution: localized adversarial training (LAT). LAT is a training tactic that focuses the classifier on certain regions in an image. Because CNNs' weights are based on every pixel of every training image, CNNs are frequently background-focused \cite{DBLP:journals/corr/RibeiroSG16}. \textit{By including adversarial backgrounds in the training set, we train the model to focus on the object more than the background}, in turn increasing the robustness of the classifier on both adversarial and natural examples.
We successfully implemented this strategy with the MNIST and CIFAR-10 datasets, creating models that outperform a traditional classifier. For the MNIST models, the localized adversarial training decreased the accuracy loss by 4.35\% on natural inputs and at least 99\% on all attempted adversarial inputs. For the CIFAR-10 models, loss decreased by 1.63\% on natural inputs and 15.98\% on adversarial inputs.

\subsection{Contributions} We first show that image classifiers may be unnecessarily dependent on image backgrounds, then develop a technique that focuses training on image foregrounds. Our contributions are as follows:
\begin{itemize}
    \item We conduct, to our knowledge, the first adversarial attack that is restricted to making limited changes to only background pixels. We run this attack against Inception\_v3 on nine images, bringing its accuracy to 0.00\%.
    \item We implement a whitebox version of this attack against an MNIST classifier, bringing its accuracy from 99.19\% to 0.08\%, and against a CIFAR-10 model, bringing its accuracy from 95.17\% to 59.26\%.
    \item We design a new, localized adversarial training technique that introduces  adversarial backgrounds into training. 
    \item We test our algorithm extensively on MNIST and CIFAR-10. In contrast to most adversarial models, which perform better on adversarial examples and worse on natural examples, the locally adversarial models performed better on both.
\end{itemize}

The rest of this paper is organized as follows: Section 2 outlines related work on adversarial attacks. Section 3 discusses our unique localized attack and benchmarks its effectiveness. In Section 4 we introduce our localized adversarial training process, and in Section 5 we implement and evaluate this training process. % creating and comparing a number of different adversarially trained models.
Section 6 discusses potential challenges for future work, and Section 7 offers our final conclusions. 

\section{Related Works} The observation that imperceptible changes could reliably fool image classifiers was first made in 2014 \cite{szegedy_zaremba_sutskever_bruna_erhan_goodfellow_fergus_2014}. More effective and reliable attacks, such as FGSM \cite{goodfellow_shlens_szegedy_2015} and PGD \cite{kurakin_goodfellow_bengio_2017}, were subsequently introduced. PGD, one of the strongest white-box attacks against CNNs \cite{madry_makelov_schmidt_tsipras_vladu_2017}, is the attack used in this paper. PGD attacks change each pixel in order to maximize the loss of the classifier, then "clip" the changes to ensure that no pixel changes more than a given distance $\epsilon$. It continues this process for a specified number of steps, each time modifying the image slightly to increase the loss of the classifier on that image. Further optimization-based attacks have been described \cite{carlini_wagner_2016}\cite{DBLP:journals/corr/LiuCLS16}, including localized adversarial attacks, or attacks where only certain pixels are altered. %(i.e. $\epsilon$ is set to $0$ at those pixels). 

Such research has explored which pixels have the biggest impact in creating adversarial examples \cite{narodytska_kasiviswanathan_2016}, how many pixels must be changed to fool a classifier \cite{papernot_mcdaniel_jha_fredrikson_celik_swami_2015}, whether a classifier can be fooled with a single pixel \cite{su_vargas_kouichi_2018} or patch \cite{brown_mane_roy_abadi_gilmer_2018}\cite{karmon_zoran_goldberg_2018},  and which pixels can be changed the most while leaving the changes imperceptible to a human \cite{luo_liu_wei_xu_2018}. The isolation of adversarial changes to a specific region allows for the testing of specific hypotheses. For instance, using a localized attack to explore the weaknesses of deep neural networks\cite{narodytska_kasiviswanathan_2016} or the transferability of attacks in a black-box setting\cite{karmon_zoran_goldberg_2018}. Localized attacks are also relevant in real-world adversarial attacks. Since an attacker can rarely alter an entire scene, physical adversarial patches \cite{brown_mane_roy_abadi_gilmer_2018}\cite{kurakin_goodfellow_bengio_2017} offer a practical means to fool an object detector or image classifier. Defenses have also been proposed for localized attacks \cite{naseer_khan_porikli_2018}. 

\section{Localized Adversarial Attacks on Background Pixels}  The localized attacks and defenses mentioned above and in Section~2 all have one of two qualities: either they focus on pixels not belonging to the object being detected, or they do not "clip" their changes to stay within a certain distance from the original image. \textit{None of the attacks which leave the main object in the image unaltered also limit the amount of noise that can be added to any other pixel. In this sense, our attack is unique.} We explore the effectiveness of attacks that are localized to the background while still attempting to restrict the perceptibility of the changes by constraining the maximum alteration of any pixel. 

\begin{figure*}
  \begin{tabular}{@{}c@{}}
    \includegraphics[scale=.23]{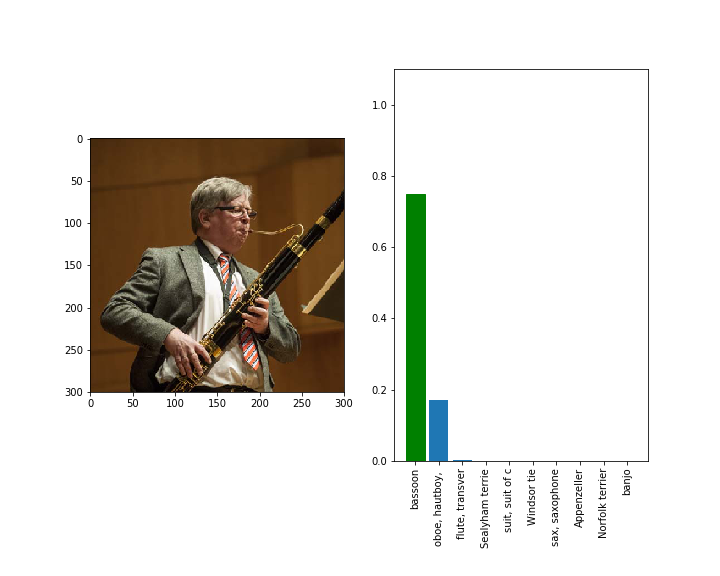} \\[\abovecaptionskip]
    \small (a.1) Classification: Bassoon (.75 confidence)
  \end{tabular}
  \begin{tabular}{@{}c@{}}
    \includegraphics[scale=.23]{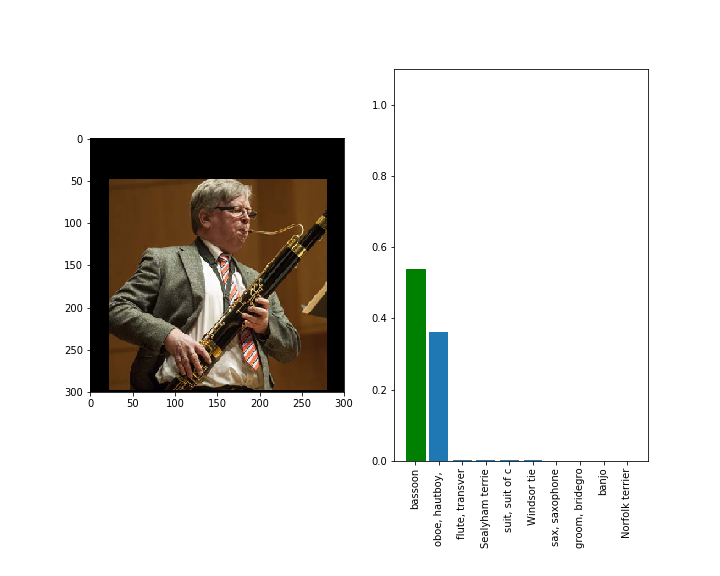} \\[\abovecaptionskip]
    \small (a.2) Bassoon (.54 confidence)
  \end{tabular}
  \begin{tabular}{@{}c@{}}
    \includegraphics[scale=.23]{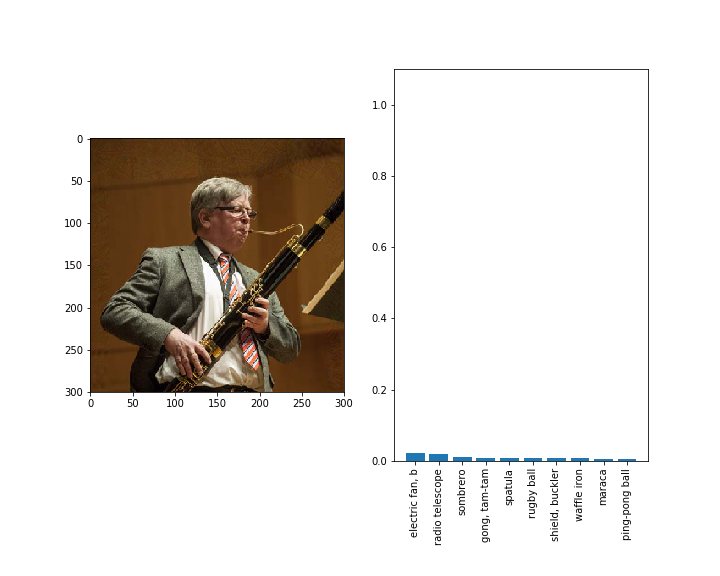} \\[\abovecaptionskip]
    \small (a.3) Electric fan (.12 confidence)
  \end{tabular}
    \begin{tabular}{@{}c@{}}
    \includegraphics[scale=.23]{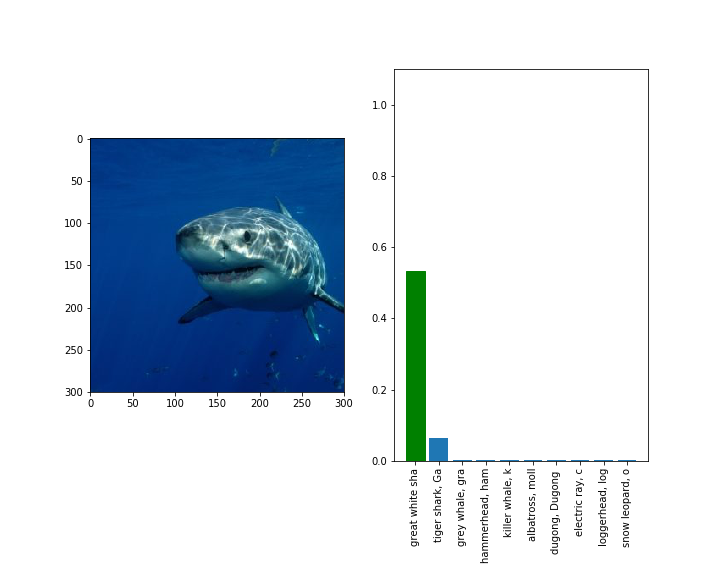} \\[\abovecaptionskip]
    \small (b.1) Classification: Great White (.53 confidence)
  \end{tabular}
  \begin{tabular}{@{}c@{}}
    \includegraphics[scale=.23]{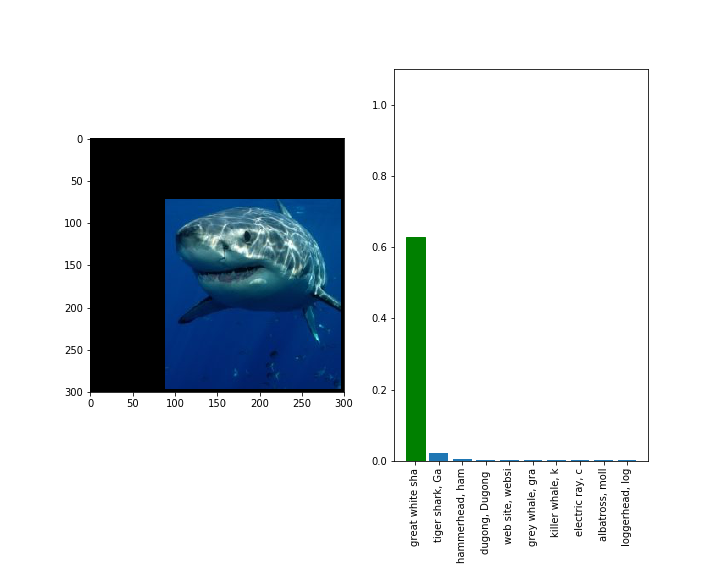} \\[\abovecaptionskip]
    \small (b.2) Great White (.63 confidence)
  \end{tabular}
  \begin{tabular}{@{}c@{}}
    \includegraphics[scale=.23]{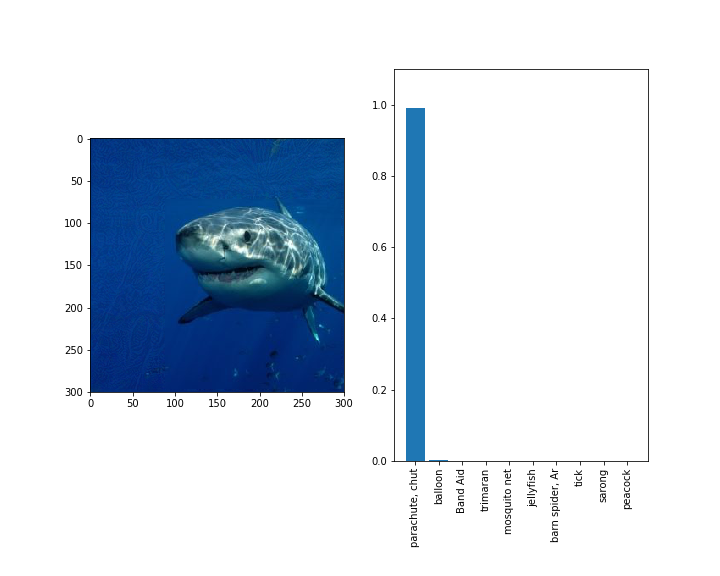} \\[\abovecaptionskip]
    \small (b.3) Parachute (.99 confidence)
  \end{tabular}
\caption{Localized and Imperceptible Adversarial Attack. The first image of each type is the unaltered, correctly identified image. The second image has any pixel outside the bounding box as black. The third, and incorrectly identified, image of each type has every pixel outside the bounding box as adversarial. The green bars represent correct guesses, and the height of each bar indicates the classifiers' confidence in that guess.}
\label{fig:advlocal}
\end{figure*}

To get a benchmark for whether or not localized and imperceptible adversarial attacks can be effective, we conduct a white-box attack on an Inception v3 image classifier \cite{DBLP:journals/corr/SzegedyVISW15}. We first use MobileNet to create a bounding box suggestion around the object, then use a PGD attack adapted from \cite{athalye} to maximize the loss of the classifier on that image. By setting $\epsilon$ to zero for every pixel inside the bounding box, we keep all pixels belonging to the object unchanged and maximize the loss function only over the remaining pixels.

We run this attack with nine 300x300 images on Inception v3, and find that the correct class was not in the top ten guesses for any of the adversarial images, suggesting that this attack is indeed effective. \textit{For these locally adversarial images, the classifier's confidence of the correct class is only .0002 on average, compared to the original confidence of .71 on the same natural inputs.} Additionally, we find that about 49.1\% of the pixels are changed. Because the bounding-boxes are imperfect, a few of those pixels don't belong to the background and therefore would not be included in a pure localized adversarial-background attack. 

A few examples of the localized attack process are included in Figure 1. The images on the left are the unaltered (and correctly identified) images, the column in the middle is used for illustrating which pixels are altered, and the column on the right holds the (incorrectly identified) localized adversarial examples, which look almost identical to the images on the left but have adversarial background pixels. The results of this experiment are summarized in Tables 1 and 2. Table 1 shows the classifier's guesses and confidences for natural images, while Table 2 shows the same classifier's guesses and confidences for the locally adversarial examples generated by our attack. Although this particular attack was run on a limited number of locally adversarial examples, we include them as an illustration and use them as a benchmark to suggest the sensitivity of classifiers to changes in the background of an image.

\begin{table}[h]
\caption{Inception v3 on unaltered inputs}
\resizebox{\linewidth}{!}{
\begin{tabular}{lllllll}
\hline
Correct Class & \begin{tabular}[c]{@{}l@{}}Percent of \\ Pixels Altered\end{tabular} & correct guess? & in top 5? & in top 10? & Classifier Guess & \begin{tabular}[c]{@{}l@{}}Confidence in \\ Correct Class\end{tabular} \\ \hline
\multicolumn{1}{|l|}{triceratops} & \multicolumn{1}{l|}{0} & \multicolumn{1}{l|}{1} & \multicolumn{1}{l|}{1} & \multicolumn{1}{l|}{1} & \multicolumn{1}{l|}{triceratops} & \multicolumn{1}{l|}{0.980953} \\ \hline
\multicolumn{1}{|l|}{tarantula} & \multicolumn{1}{l|}{0} & \multicolumn{1}{l|}{1} & \multicolumn{1}{l|}{1} & \multicolumn{1}{l|}{1} & \multicolumn{1}{l|}{tarantula} & \multicolumn{1}{l|}{0.957682} \\ \hline
\multicolumn{1}{|l|}{snail} & \multicolumn{1}{l|}{0} & \multicolumn{1}{l|}{1} & \multicolumn{1}{l|}{1} & \multicolumn{1}{l|}{1} & \multicolumn{1}{l|}{snail} & \multicolumn{1}{l|}{0.880702} \\ \hline
\multicolumn{1}{|l|}{bassoon} & \multicolumn{1}{l|}{0} & \multicolumn{1}{l|}{1} & \multicolumn{1}{l|}{1} & \multicolumn{1}{l|}{1} & \multicolumn{1}{l|}{bassoon} & \multicolumn{1}{l|}{0.748791} \\ \hline
\multicolumn{1}{|l|}{hammer} & \multicolumn{1}{l|}{0} & \multicolumn{1}{l|}{1} & \multicolumn{1}{l|}{1} & \multicolumn{1}{l|}{1} & \multicolumn{1}{l|}{hammer} & \multicolumn{1}{l|}{0.925929} \\ \hline
\multicolumn{1}{|l|}{iPod} & \multicolumn{1}{l|}{0} & \multicolumn{1}{l|}{1} & \multicolumn{1}{l|}{1} & \multicolumn{1}{l|}{1} & \multicolumn{1}{l|}{iPod} & \multicolumn{1}{l|}{0.982066} \\ \hline
\multicolumn{1}{|l|}{great white shark} & \multicolumn{1}{l|}{0} & \multicolumn{1}{l|}{1} & \multicolumn{1}{l|}{1} & \multicolumn{1}{l|}{1} & \multicolumn{1}{l|}{great white} & \multicolumn{1}{l|}{0.53433} \\ \hline
\multicolumn{1}{|l|}{hen} & \multicolumn{1}{l|}{0} & \multicolumn{1}{l|}{1} & \multicolumn{1}{l|}{1} & \multicolumn{1}{l|}{1} & \multicolumn{1}{l|}{hen} & \multicolumn{1}{l|}{0.380655} \\ \hline
\multicolumn{1}{|l|}{kite} & \multicolumn{1}{l|}{0} & \multicolumn{1}{l|}{0} & \multicolumn{1}{l|}{0} & \multicolumn{1}{l|}{0} & \multicolumn{1}{l|}{umbrella} & \multicolumn{1}{l|}{0.00108} \\ \hline
Average & 0 & 0.888889 & 0.888889 & 0.888889 &  & 0.710243 \\ \hline
\end{tabular}}
\end{table}

\begin{table}[h]
\caption{Inception v3 on locally adversarial images}
\resizebox{\linewidth}{!}{
\begin{tabular}{lllllll}
\hline
Correct Class & \begin{tabular}[c]{@{}l@{}}Percent of \\ Pixels Altered\end{tabular} & correct guess? & in top 5? & in top 10? & Classifier Guess & \begin{tabular}[c]{@{}l@{}}Confidence in \\ Correct Class\end{tabular} \\ \hline
\multicolumn{1}{|l|}{triceratops} & \multicolumn{1}{l|}{35.6} & \multicolumn{1}{l|}{0} & \multicolumn{1}{l|}{0} & \multicolumn{1}{l|}{0} & \multicolumn{1}{l|}{coral reef} & \multicolumn{1}{l|}{0.000377} \\ \hline
\multicolumn{1}{|l|}{tarantula} & \multicolumn{1}{l|}{61} & \multicolumn{1}{l|}{0} & \multicolumn{1}{l|}{0} & \multicolumn{1}{l|}{0} & \multicolumn{1}{l|}{wolf spider} & \multicolumn{1}{l|}{4.44E-08} \\ \hline
\multicolumn{1}{|l|}{snail} & \multicolumn{1}{l|}{70.7} & \multicolumn{1}{l|}{0} & \multicolumn{1}{l|}{0} & \multicolumn{1}{l|}{0} & \multicolumn{1}{l|}{bakery} & \multicolumn{1}{l|}{2.42E-06} \\ \hline
\multicolumn{1}{|l|}{bassoon} & \multicolumn{1}{l|}{28.3} & \multicolumn{1}{l|}{0} & \multicolumn{1}{l|}{0} & \multicolumn{1}{l|}{0} & \multicolumn{1}{l|}{electric fan} & \multicolumn{1}{l|}{0.000657} \\ \hline
\multicolumn{1}{|l|}{hammer} & \multicolumn{1}{l|}{15.5} & \multicolumn{1}{l|}{0} & \multicolumn{1}{l|}{0} & \multicolumn{1}{l|}{0} & \multicolumn{1}{l|}{hand blower} & \multicolumn{1}{l|}{0.000389} \\ \hline
\multicolumn{1}{|l|}{iPod} & \multicolumn{1}{l|}{70.3} & \multicolumn{1}{l|}{0} & \multicolumn{1}{l|}{0} & \multicolumn{1}{l|}{0} & \multicolumn{1}{l|}{paper towel} & \multicolumn{1}{l|}{1.10E-07} \\ \hline
\multicolumn{1}{|l|}{great white shark} & \multicolumn{1}{l|}{48} & \multicolumn{1}{l|}{0} & \multicolumn{1}{l|}{0} & \multicolumn{1}{l|}{0} & \multicolumn{1}{l|}{parachute} & \multicolumn{1}{l|}{1.10E-07} \\ \hline
\multicolumn{1}{|l|}{hen} & \multicolumn{1}{l|}{36.3} & \multicolumn{1}{l|}{0} & \multicolumn{1}{l|}{0} & \multicolumn{1}{l|}{0} & \multicolumn{1}{l|}{potter's wheel} & \multicolumn{1}{l|}{9.68E-05} \\ \hline
\multicolumn{1}{|l|}{kite} & \multicolumn{1}{l|}{76} & \multicolumn{1}{l|}{0} & \multicolumn{1}{l|}{0} & \multicolumn{1}{l|}{0} & \multicolumn{1}{l|}{brain coral} & \multicolumn{1}{l|}{6.02E-11} \\ \hline
Average & 49.07778 & 0 & 0 & 0 &  & 0.000169 \\ \hline
\end{tabular}}
\end{table}

\section{Localized Adversarial Training} 
Because adversarial attacks expose such a glaring weakness in traditional image classifiers, it is prudent to judge the quality of a classifier not simply on its accuracy on natural inputs, but also on its robustness when faced with adversarial inputs. One particular method of achieving robustness for a classifier is adversarial training. In this section, we introduce our new localized adversarial training algorithm, which trains our classifier to focus on the foreground. 

As mentioned in Section 1, a degree of robustness (as well as extra regularization \cite{goodfellow_shlens_szegedy_2015}) can be achieved by including adversarial examples in the training process. Because a classifier's weights are continually updated during training, adversarial training requires that newly generated adversarial examples be introduced at every step. This slowly trains the model to perceive adversarial examples and natural examples the same way, making it more resistant to unseen adversarial examples. However, despite the robustness of adversarially trained models, they are generally outperformed on natural inputs by their non-adversarially trained counterparts. This tradeoff is explained in \cite{tsipras_santurkar_engstrom_turner_madry_2018} as a result of the fact that the most accurate models rely on "non-robust features," or features that are most subject to adversarial attacks. It is here that our localized attacks become most relevant: {if a model is adversarially trained to discount \textit{irrelevant} features (like the background, in many cases), then it may become more robust to adversarial changes in those features without suffering the same accuracy loss of a standard adversarially-trained model, which is trained to not rely heavily on \textit{any} feature.} The details of this procedure are found in Algorithm 1.

As described in Algorithm 1, LAT iterates through small batches of the training data, making each image in each batch adversarial. For each image, a PGD attack generates and adds adversarial perturbations. Then, the attack is localized by creating a matrix of equal size to the image, denoted as $epmatrix$, where each value is $\epsilon$ if the corresponding pixel is altered, or zero otherwise. (In the next Section, we will describe in detail the different localized $attacks$ we explored.) The adversarial image is then clipped, to ensure that its distance from the original image at any given pixel is no more than that pixel's corresponding value in the epsilon matrix (meaning that higher values of $\epsilon$ allow more visible changes at those pixels). Finally, once all the images in a batch are adversarial, the neural network is updated and trained to recognize those images correctly.

\begin{algorithm}
\caption{Localized Adversarial Training}
\begin{algorithmic}[1]
\State $n$ is a CNN; $\epsilon$ is the maximum value that any pixel may legally change; $attack$ describes which pixels may legally change
\Repeat \ for each minibatch \textit{B} in training data
\Repeat \ for each image in \textit{B}
    \State $x\gets image$
    \State $\lambda\gets $ {PGD\_Attack ($x, n$)} \Comment{Noise generated by PGD attack}
    \State $x'\gets x + \lambda$
    \State $epmatrix$ is initialized  
    \State $epmatrix \gets $Localize ($epmatrix, \epsilon, attack$) \Comment{Changes are localized}  
    \State $x'$ clipped to within $x$+$epmatrix$, $x$-$epmatrix$
    \State replace original image in \textit{B} with $x'$
\Until every image in batch is altered
\State Train \textit{n} on updated batch \textit{B}
\Until training is complete
\end{algorithmic}
\end{algorithm}

\section{Evaluation of Robustness and Accuracy} To test the robustness and accuracy of locally adversarially trained models, we train 3 CNN classifiers each for the CIFAR-10 and MNIST datasets. Each model is trained with two convolutional layers, a fully connected layer, and an output layer, and undergoes 100,000 steps of training. The first model for each dataset is a "natural" model, which is trained on unaltered images. The "standard" model is trained on standard, traditional adversarial images, with every pixel altered. Finally, the "background" model is trained on locally adversarial images with the middle pixels unaltered. Each attack is iterated over 100 steps, and $\epsilon$ is set to .3 for every pixel that is allowed to change.

\subsection{Experiment Results} 
  Evaluating each of the three models on each of the three inputs types yields a total of 9 results per model. The experiment confirms the trade-off between accuracy and robustness for standard adversarial training: while the MNIST model trained with the standard attack suffered at least 99\% less loss than the naturally trained model, it suffered 39.9\% more loss on natural inputs. Similarly, while the CIFAR-10 standard adversarial model suffered 95.7\% less loss on adversarial images, it suffered 88.9\% more loss on natural images.
  
  %\textit{Importantly, this trade-off did not apply to the models trained on localized adversarial examples.} 
  For both datasets, the "background" models outperformed the natural models, by 4.35\% on natural MNIST inputs and over 99\% on adversarial MNIST inputs. The "background" CIFAR-10 model also had improved performance in both categories, by 1.63\% on natural inputs and 15.98\% on adversarial inputs. Figure 2 describes the loss of each model when tested on natural inputs and each kind of adversarial input. Only the models which underwent LAT outperformed the "natural" model on all inputs.

\begin{figure}[h]
\centering
\includegraphics[scale=.5]{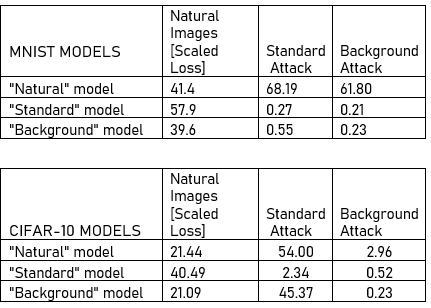}
\caption{The cross entropy loss of each model on natural inputs and each type of adversarial input. Loss on natural inputs is scaled by a factor of 1000 for MNIST and 100 for CIFAR-10.}
\label{fig:results}
\end{figure}

\section{Discussion} We show that an MNIST and CIFAR-10 classifier can be improved with localized adversarial training. We also show the effectiveness of localized adversarial attacks against complex, high-resolution image classifiers like Inception v3. Although we did not use LAT for more complex, higher-resolution image classifiers, our success with LAT on the MNIST and CIFAR-10 classifiers, as well as with the localized adversarial attack against Inception v3, indicates that LAT is a promising next step to increase the accuracy and robustness of those classifiers as well. Because LAT can be implemented as a modification to preexisting adversarial training techniques, the main challenge for LAT is identifying which pixels should be changed. While it is computationally inexpensive to use semantic segmentation or foreground/background segregation on a low-resolution, black-and-white dataset like MNIST, it is much more computationally expensive to identify background pixels on high-resolution color images. A much cheaper approach is to leave the middle pixels unaltered, which, although effective in our experiment, may not yield the same benefits as a more exact masking approach when used on complex, high-resolution classification. Still, either strategy could still lead to improved accuracy on natural inputs compared to traditional adversarial training. Finally, we suggest that LAT may be an effective tool to combat bias in deep learning. We used it to prevent image classifiers from focusing on the background, but it could just as easily be used to prevent a risk-calculator from focusing on elements like race or skin-color. We leave these applications for future work.

\section{Conclusion} We conclude that CNN classifiers are often too sensitive to changes in the background, and that this can be addressed by including images with adversarial backgrounds in the training set. This focuses the training on the image foregrounds, increasing accuracy and robustness. Localized adversarial training is cheap to implement and could have broad applications.

\section{Acknowledgement} This work was supported in part by NSF grant CNS-1758017.

\bibliographystyle{ACM-Reference-Format}
\bibliography{references}

\end{document}